\documentclass{article}

\usepackage{arxiv}

\usepackage[utf8]{inputenc} % allow utf-8 input
\usepackage[T1]{fontenc}    % use 8-bit T1 fonts
\usepackage{hyperref}       % hyperlinks
\usepackage{url}            % simple URL typesetting
\usepackage{booktabs}       % professional-quality tables
\usepackage{amsmath,amssymb,amsfonts}
\usepackage{amsfonts}       % blackboard math symbols
\usepackage{nicefrac}       % compact symbols for 1/2, etc.
\usepackage{microtype}      % microtypography
\usepackage{cleveref}       % smart cross-referencing
\usepackage{lipsum}         % Can be removed after putting your text content
\usepackage{graphicx}
\usepackage{natbib}
\usepackage{doi}

\usepackage{mathtools}
\usepackage{balance}
\usepackage{subcaption}

\usepackage{tikz}
\usetikzlibrary{positioning, shapes.multipart}
\usepackage{pgfplots}
\pgfplotsset{compat=1.18}
\usetikzlibrary{calc}
\graphicspath{ {./images/} }
\usepackage{lineno}
\usepackage[final]{changes}
\usepackage{comment}

\newcommand{\bftab}{\fontseries{b}\selectfont}

\DeclareMathOperator*{\argmin}{argmin}

\title{Quantification via Gaussian Latent Space Representations }

\author{ \href{https://orcid.org/0000-0002-4527-6698}{\includegraphics[scale=0.06]{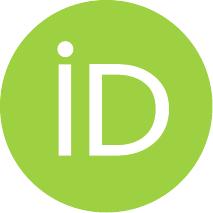}\hspace{1mm}Olaya Pérez-Mon}\\
	Artificial Intelligence Center\\
	University of Oviedo\\
	Gijón, 33204, Asturias, Spain \\
	\texttt{perezolaya@uniovi.es} \\
	\And
	\href{https://orcid.org/0000-0002-0377-1025}{\includegraphics[scale=0.06]{orcid.pdf}\hspace{1mm}Juan José del Coz} \\
	Artificial Intelligence Center\\
	University of Oviedo\\
	Gijón, 33204, Asturias, Spain \\
	\texttt{juanjo@uniovi.es} \\
        \And
	\href{https://orcid.org/0000-0002-9250-0920}{\includegraphics[scale=0.06]{orcid.pdf}\hspace{1mm}Pablo González} \\
	Artificial Intelligence Center\\
	University of Oviedo\\
	Gijón, 33204, Asturias, Spain \\
	\texttt{gonzalezgpablo@uniovi.es} \\
}

\begin{document}
\maketitle
\begin{abstract}
\noindent Quantification, or prevalence estimation, is the task of predicting the prevalence of each class within an unknown bag of examples. Most existing quantification methods in the literature rely on prior probability shift assumptions to create a \added{quantification} model that uses the predictions of an underlying classifier to make optimal prevalence estimates. In this work, we present an end-to-end neural network that uses Gaussian distributions in latent spaces to obtain invariant representations of bags of examples. This approach addresses the quantification problem using deep learning, enabling the optimization of specific loss functions relevant to the problem and avoiding the need for an intermediate classifier, tackling the quantification problem as a direct optimization problem. Our method achieves state-of-the-art results, both against traditional quantification methods and other deep learning approaches for quantification. The code needed to reproduce all our experiments is publicly available at \url{https://github.com/AICGijon/gmnet}.\end{abstract}

% keywords can be removed
%\keywords{First keyword \and Second keyword \and More}

\section{Introduction}

Across a wide range of machine learning tasks, deep learning has proven a very useful, even a revolutionary tool to obtain amazing results in a wide variety of applications. These include generative IA (text, image, video), translation, or computer vision, among others. Despite its unprecedented success in these areas, the application of deep learning techniques in more traditional machine learning tasks, such as quantification, has been less straightforward. 

Quantification, or prevalence estimation, involves the challenge of developing models capable of accurately estimating the prevalence of different classes within sets of examples unseen during training. These sets, often referred to as `bags', are affected by some kind of dataset shift (typically prior probability shift), which makes the problem non-trivial. In the field of quantification, there is a variety of machine learning methods designed to tackle this problem \citep{gonzalez2017review, quantbook2023}. Among these methods, the most successful ones (which will be introduced in the following section) typically employ an underlying classifier as the foundation upon which the quantification method is built \citep{schumacher2021comparative}. 

A key aspect of these quantification methods is their learning assumption. Under prior probability assumptions \citep{MorenoTorres2012}, the distribution of data varies between train and test, characterized by the condition $P_{tr}(Y) \neq P_{tst}(Y)$ while maintaining $P_{tr}(X|Y) = P_{tst}(X|Y)$. In other words, the prevalence of the classes changes between train and test, but the class conditionals remain constant. In particular, the assumption $P_{tr}(Y) \neq P_{tst}(Y)$ makes that the obvious approach to solving quantification, that is 1) train a classifier using training data; 2) classify the examples in the test bag; and 3) count how many belong to each class, does not work in practice. This method is named Classify \& Count (CC) in the quantification literature and is not suited for quantification problems because the classifier's output probabilities are calibrated based on the training distribution. When the class distribution changes in the test bags, these probabilities no longer reflect the true prevalence of classes, leading to biased and inaccurate prevalence estimates. Specifically, this discrepancy may cause the method to \added{overestimate/underestimate the prevalence of a given class when its true prevalence decreases/increases compared to that observed in the training data.}

Recent advancements in quantification have introduced deep learning methods that diverge from classical approaches \citep{Esuli2018,Qi2021,perez2024quantification}. Unlike traditional methods, some of these deep learning techniques operate without explicit learning assumptions and can directly utilize bags of examples without the need for individual example annotations. Another distinguishing feature is their capability to optimize a user-defined loss function, tailored to the specific requirements of the quantification tasks. A fundamental characteristic of deep learning techniques applied to quantification is their focus on deriving a representative embedding for each bag of examples, crucial for estimating the bag class prevalences (the final objective). This bag representation serves as a compressed representation of the entire bag and should have the necessary information to facilitate the reconstruction of class prevalences present within the bag. The use of bags annotated by prevalence directly as input data, transforms the learning task from an \textit{asymmetric} one to a \textit{symmetric} one \citep{perez2024quantification}. In the traditional \textit{asymmetric} approach, a model is trained on individually labeled examples to predict bag prevalences, typically relying on an intermediate classifier. By contrast, the \textit{symmetric} approach addresses the quantification task more directly, using bags annotated by prevalence as input. This eliminates the need for classification as an intermediate step, allowing the model to focus directly on the \added{quantification problem.  }

In this work, we propose a new layer for representing bags of examples, based on modeling latent spaces using Gaussian distributions. The idea is to improve previous representation layers, such as pooling layers \citep{Qi2021} or histograms \citep{perez2024quantification}. We prove through experimentation that this bag representation is more powerful than the previous ones, obtaining unprecedented results in public datasets included in the only two quantification competitions held so far. Apart from the layer's inner workings and its parameter initialization, we also discuss a regularization technique, specific to this type of layer, that may help improve the results and the convergence time. \added{Finally, we perform an extensive experimentation, comparing traditional quantification methods and deep learning methods, under various settings, to study the advantages and limitations of both.}

The rest of the paper is organized as follows. Section~\ref{sec:related} introduces classical learning methods in the quantification literature. 
Then, in Section~\ref{sec:deeplearning}, we discuss previous approaches using deep learning for tackling the quantification problem, proposing our own layer for representing bags, discussing parameter initialization, regularization, and the use of example labels if available. Section~\ref{sec:experiments} details our experimental setup and findings. Finally, Section~\ref{sec:conclusions} presents our conclusions and outlines avenues for future research.

\section{Related work: previous work on quantification}
\label{sec:related}
In this section, we are going to briefly present the most commonly used quantification methods in the literature. We refer the reader to \citet{gonzalez2017review, quantbook2023} for a more in-depth explanation of these methods. 

Given a labeled training set $\mathcal{D}_{\text{tr}} = \{(\mathbf{x}_i, y_i)\}_{i=1}^{n}$, drawn from a distribution $P_{\text{tr}}(\mathcal{X}, \mathcal{Y})$ where $\mathcal{X}$ represents the feature space and $\mathcal{Y}$ the set of true class labels $\{c_1, c_2, \dots, c_l\}$, and an unlabeled test bag $B = \{\mathbf{x}_j\}_{j=1}^{m}$ drawn from a different distribution $P_{\text{tst}}(\mathcal{X}, \mathcal{Y})$ where $P_{\text{tr}}(\mathcal{Y}) \neq P_{\text{tst}}(\mathcal{Y})$, the goal is to estimate the prevalence (or proportion) $\hat{P}(\mathcal{Y})$ of each class in the test bag, assuming that $P_{tr}(\mathcal{X}|\mathcal{Y}) = P_{tst}(\mathcal{X}|\mathcal{Y})$. 

One of the first methods in the quantification literature is the Adjusted Classify and Count (ACC) method (see~\citet{Forman2008}), also known as Black Box Shift Estimation (BBSE) \citep{lipton2018detecting}. It learns a classifier \added{$h:\mathcal{X}\rightarrow \mathcal{Y}$} from training data and then applies a correction. First, we compute the
probability of predicting that a random example $x \in B$ is classified as class $c_i$:
%
% \begin{align}
%  \label{eq:ACC} 
%  \Pr(h(\mathbf{x})=y_{i}) = \sum_{y_{j}\in
%  Y\Pr(h(\mathbf{x})=y_{i}|y_{j})\cdot \Pr(y_{j})
%\end{align}
\begin{align}
  \label{eq:AC} 
  p(h(\mathbf{x})=c_{i}) = \sum_{c_{j}\in
  \mathcal{Y}}p(h(\mathbf{x})=c_{i}|y=c_{j})\cdot p(c_{j}),
\end{align}
where $p(h(\mathbf{x})=c_{i}|y=c_{j})$ is the probability that $h$ predicts $c_i$ when the actual class of $x$ is $c_j$ and $p(c_j)$ is the true prevalence of class $c_j$ in the test bag. The former probability can be estimated using many-fold cross-validation over the train set, while the latter is the quantity that we try to find out. Writing this same equation for all the classes, we get a system of $l$ equations that we can solve obtaining the values $[p(c_1), p(c_2), \dots, p(c_l)]$ which will be the model predictions for the bag class prevalences $[\hat{p}_1, \hat{p}_2, \dots, \hat{p}_l]$.

In \citet{Bella2010}, the authors propose probabilistic versions of the methods Classify and Count (CC) and Adjusted Classify and Count (ACC). Both methods replace the hard classifier $h$ with a soft classifier \added{$s:\mathcal{X}\rightarrow {[0,1]}^l$}. The former, named Probabilistic Classify and Count (PCC), just replaces basic class counting by averaging the probabilistic outputs of the classifier by class: 

\added{
\begin{equation}
\hat{p_l}=\frac{1}{m}\sum_{\mathbf{x}_j \in B} s(\mathbf{x}_j)[c_l].
\end{equation}
}

On the other hand, ACC is transformed into Probabilistic Adjusted Classify and Count (PACC), and solves a system equivalent to the one used in ACC but based on the computation of posterior probabilities averaged over $D_{tr}$ and $B$. 

Another family of quantification methods is based on matching a mixture of the training distribution and the testing distribution \citep{Maletzke2019}, based on estimates from $D_{tr}$ and $B$ respectively. The training distribution mixture $D_{tr}'$ is modeled as a combination of class-specific distributions of each class $c_j$, $D_{tr}^{c_j}$, weighted by their respective estimated prevalence $\hat{p}_j$:
\begin{equation}
D_{tr}' = \sum_{c_j \in \mathcal{Y}} D_{tr}^{c_j} \cdot \hat{p}_j.
\end{equation}
The idea is to minimize a distance, $\Delta$, between $D'_{tr}$ and the test bag distribution $B'$:
\begin{equation}
\label{eq:distributionmatching}
\argmin_{\hat{p}_1, \ldots, \hat{p}_l} \Delta(D'_{tr}, B') = \argmin_{\hat{p}_1, \ldots, \hat{p}_l} \Delta \bigg(\sum_{c_j \in \mathcal{Y}} D_{tr}^{c_j} \cdot \hat{p}_j, B'\bigg).
\end{equation}

Quantification algorithms based on matching distributions consist of three different elements: 1) a method to estimate data distributions, 2) a similarity measure $\Delta$, and 3) an optimization method to solve Equation~\ref{eq:distributionmatching} \citep{castano2023equivalence}. One popular choice is training a probabilistic classifier $s$ and representing the distributions using histograms over the example predictions \citep{GonzalezCastro2013}. In the experiments, for the similarity measure \added{$\Delta$}, we have selected the Hellinger distance \citep{GonzalezCastro2013}. We refer to this method as DMy in the results section.

The Expectation Maximization for Quantification (EMQ) method \citep{Saerens2002} uses the EM algorithm to adjust the posterior probabilities produced by a soft classifier $s$ in response to potential shifts in label distribution. This process involves iterating through mutually recursive steps: expectation (updating the posteriors) and maximization (updating the priors) until convergence is achieved. This method is among the most competitive in quantification literature \citep{lequa2022}, but it heavily relies on the calibration of the posterior probabilities returned by the soft classifier $s$ \citep{alexandari2020maximum}.

\section{Tackling quantification using deep learning}
\label{sec:deeplearning}
Quantification was historically tackled with machine learning algorithms that worked under the prior probability shift conditions (see Section~\ref{sec:related}). Usually, these methods worked using the predictions of a classifier trained over individual examples. Recently, deep learning techniques have allowed to reformulate the task into an end-to-end quantification task \citep{perez2024quantification}. 

The difference between the two approaches is notable: traditional methods use an \textit{asymmetric} approach, requiring a labeled training set $\mathcal{D}_{\text{tr}} = \{(\mathbf{x}_i, y_i)\}_{i=1}^{n}$, where each individual example $\mathbf{x}_i$ is associated with a label $y_i \in \mathcal{Y}$. In this approach, a classifier $h: \mathcal{X} \to \mathcal{Y}$ \added{or $s:\mathcal{X} \rightarrow {[0,1]}^l$} is trained on $\mathcal{D}_{\text{tr}}$ to predict labels \added{or posterior probabilities} for individual examples, and the quantifier is derived by aggregating these predictions over a test bag $B = \{\mathbf{x}_j\}_{j=1}^{m}$, often assuming prior probability shift between training and test data. This approach does not directly minimize a quantification-specific loss and instead relies on classification as an intermediate step.

In contrast, deep learning methods \citep{Qi2021,perez2024quantification} adopt a \textit{symmetric} approach, where the model is trained directly on bags of examples $\mathcal{D}_{\text{tr}} = {\{(B_i, \mathbf{p}_i)\}}_{i=1}^{n}$, with each training bag $B_i = {\{\mathbf{x}_{j}\}}_{j=1}^{m}$ labeled by its prevalence vector $\mathbf{p}_i = [p_i(c_1), \dots, p_i(c_l)]$, where $p_i(c_k)$ is the proportion of class $c_k$ in the bag $i$. During training, the network weights are optimized to predict the prevalence vector $\mathbf{p}_i$ for each training bag, by directly minimizing a quantification-specific loss, such as mean squared error or another user-defined error measure. This eliminates the need for intermediate classification and allows the model to focus on learning group-level patterns, providing more accurate and robust prevalence estimates. \added{Notice also that this approach does not assume a particular shift (prior probability shift, covariate shift, etc); in fact, the training bags may present any shift.}

\subsection{Previous deep learning approaches used in quantification}

In this section, we will summarize previous deep learning approaches developed to address the quantification problem.

The first approach, called QuaNet \citep{Esuli2018}, introduces an innovative method using a Recurrent Neural Network (RNN) for binary sentiment quantification. This involves estimating the relative frequency of positive and negative sentiment labels in a collection of texts. QuaNet utilizes the predictions from an external probabilistic classifier $s$, which are sorted and fed into the network. This enables the RNN module to learn to identify the switching point between the positive and negative classes. Additionally, prevalence predictions computed using traditional quantifiers like CC, PCC, ACC, and PACC are concatenated in the last layers of the network to enhance the representation obtained by the RNN. Even though QuaNet was the first deep neural network designed to tackle the quantification problem, it has some major drawbacks. First, it needs an underlying classifier, which is needed to obtain the traditional quantifiers predictions, which means that the method needs example labels to be trained. Second and more important, it only works for binary quantification, as its concept of a cut point relies on having just two classes in the dataset. This limitation makes it unsuitable for this study as we are dealing with problems with more than two classes.

The second attempt to solve quantification using deep learning was proposed by \citet{Qi2021} and it is called Deep Quantification Network (DQN). This paper proposes a deep learning architecture to tackle quantification using a \textit{symmetric} approach. The architecture consists of three main components:
\begin{enumerate}
    \item \textbf{Feature Extraction Module (FEM)}: This module's layers vary based on the specific data and may include fully connected layers, Convolutional Neural Networks (CNNs), transformers, etc. It is in charge of computing feature vectors for individual examples, applying transformations to the examples, and projecting them to a latent space.

    \item \textbf{Bag Representation Module (BRM)}: This module summarizes all the feature vectors of the examples in a bag into a single feature vector representing the entire bag. The BRM in DQN utilizes basic pooling layers, such as \textit{average}, \textit{max}, or \textit{median}, which are permutation invariant to the order of the examples within the bag.

    \item \textbf{Quantification Module (QM)}: This module connects the bag representation to a final layer that predicts the prevalence of each class. The QM comprises a set of fully connected layers with a final softmax activation function to ensure that the predicted prevalences sum to one.
\end{enumerate}

It is important to note that the approach proposed by \citet{Qi2021}, was previously proposed by \citet{zaheer2017deep}, just not for quantification purposes but for set processing in general. In their paper, the authors emphasize the importance of a permutation invariant representation when dealing with sets. The underlying intuition is straightforward: the feature vector computed for the set should remain the same regardless of the order of the examples within the set. This characteristic is evidently applicable to the quantification task as well.

The main advantage of the DQN architecture is that it offers a \textit{symmetric} solution to the quantification problem while also being capable of optimizing a specific loss function. However, the primary drawback is that the basic pooling layers used in the BRM in DQN are not sufficiently representative to achieve state-of-the-art quantification results and could benefit from further improvement.

A step further in this direction was the application of differentiable histograms proposed by \citet{perez2024quantification}, giving place to a network called HistNetQ. In this work, authors propose a BRM that summarizes the bag computing a histogram for each feature computed by the FEM. \added{This histogram layer represented a generalization of the simple pooling layers used in DQN, providing the BRM with additional information useful for quantification tasks.} Since histograms are inherently non-differentiable, the authors propose to use a differentiable approximation using typical deep learning layers as convolutions. This new BRM, along with some innovative techniques for using data augmentation in quantification tasks \added{to avoid overfitting when training bags are scarce (see Section~\ref{sec:dataaugmentation})}, got very good results in the 2022 quantification competition \citep{lequa2022}, in particular in the multiclass quantification task (T1B), obtaining the best results so far in the quantification literature \citep{perez2024quantification}.

In the next section, we propose a new BRM based on the use of Gaussian distributions for modeling the latent space computed by the FEM.

\begin{figure}[t]
\centering
        \includegraphics[width=\textwidth, trim=0 30 0 0, clip]{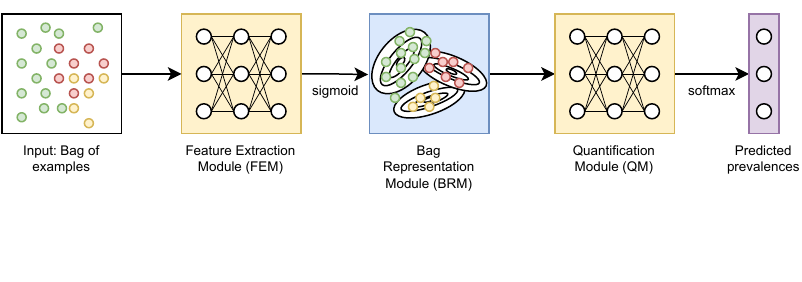}
        \caption{Basic network architecture for a problem with three classes. The network processes a bag of examples and projects it into a latent space through a Feature Extraction Module (FEM). Then, an order invariant representation of the bag is computed by the Bag Representation Module (BRM). In the figure, Gaussian distributions are used for obtaining the bag representation (as explained in Section~\ref{sec:latentrepr}). Other permutation invariant layers as basic pooling layers or differentiable histograms might be used instead to represent the bags. Finally, the Quantification Module (QM), composed by a set of fully connected layers, relates the bag representation vector to the actual class prevalences. Note that the FEM can be adapted to the problem at hand (CNN for images, Transformers for text, etc).}
        \label{fig:architecture}
\end{figure}

\subsection{Representing sets using Gaussian distributions in a latent space}
\label{sec:latentrepr}

%In previous quantification networks, achieving invariant representations involved various methodologies. DQN \citep{Qi2021} utilized basic pooling layers such as \textit{average},  \textit{median}, or \textit{max} to compute invariant representations of bags in the latent space. HistNetQ \citep{perez2024quantification} advanced this approach by employing differentiable histograms of latent space features, demonstrating enhanced information capture suitable for the quantification task, compared to basic pooling layers.

In this study, we propose a novel method for obtaining invariant bag features using learnable multivariate Gaussian distributions. The core idea behind this representation layer is to enable the network to model the latent space using a set of multivariate Gaussian distributions where each example's projection in the latent space (bounded in a hypercube of size 1 by a sigmoid activation function) is evaluated against these distributions.

The intuition behind using Gaussians is that they capture complex, multidimensional relationships between features simultaneously, rather than focusing on individual features, simple aggregate statistics \citep{Qi2021} or differentiable histograms \citep{perez2024quantification}. By modeling the latent space with Gaussians, the network can identify distinct regions within the space where examples are likely to cluster based on feature patterns relevant to quantification. This helps create a more nuanced, continuous representation of the data that is adaptable to shifts in prevalence, ultimately leading to more accurate prevalence estimates.

Let \( \mathbf{Z} = [\mathbf{z}_1, \mathbf{z}_2, \ldots, \mathbf{z}_n] \) denote the latent space representations of \( n \) examples in a training bag, where each \( \mathbf{z}_i \in \mathbb{R}^d \) represents a vector of \( d \) features (latent space dimension). Suppose we model the latent space using \( K \) Gaussian distributions where \( \boldsymbol{\mu}_k \) and \( \boldsymbol{\Sigma}_k \) are network learnable parameters representing the mean vector and covariance matrix of the \( k \)-th Gaussian distribution.

\begin{figure}[t]
\centering
        \includegraphics[width=0.8\textwidth]{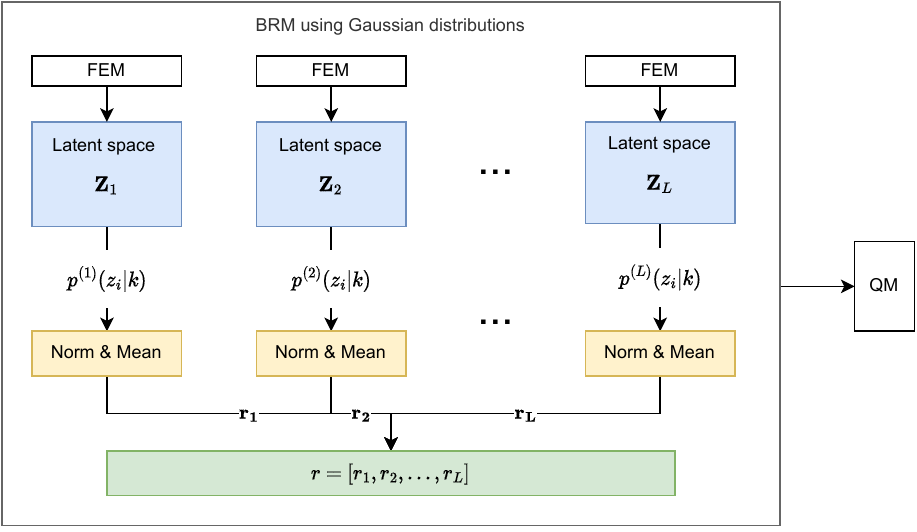}
        \caption{Extension for using multiple latent spaces representations. Each latent space $\mathbf{Z}_l$ can have a different number of dimensions $d$, and a different number $K$ of Gaussian distributions to model it (hyperparameters). After the examples in the bag are projected to each latent space, we compute the likelihood $p^{(l)}(\mathbf{z}_i|k)$ of each example $i$ to belong to each Gaussian distribution $k=1..K$ in a latent space $l=1..L$. Then we apply normalization and the mean by Gaussian, getting a vector $\mathbf{r}_l$. The final representation layer is the concatenation of each vector $\mathbf{r}_l$. }
        \label{fig:architecturehorizontal}
\end{figure}

For each Gaussian \( k \), the likelihood \( p(\mathbf{z}_i | k) \) for example \( i \) represents how well \( \mathbf{z}_i \) fits Gaussian \( k \):
\begin{equation}
p(\mathbf{z}_i | k) = \frac{1}{\sqrt{(2\pi)^{d} |\boldsymbol{\Sigma}_k|}} \exp \left( -\frac{1}{2} (\mathbf{z}_i - \boldsymbol{\mu}_k)^T \boldsymbol{\Sigma}_k^{-1} (\mathbf{z}_i - \boldsymbol{\mu}_k) \right).
\end{equation}
To obtain the mean likelihood across all examples in the bag for Gaussian \( k \), denoted by \( \bar{p}(k) \), we compute:
\begin{equation}
\bar{p}(k) = \frac{1}{n} \sum_{i=1}^{n} p(\mathbf{z}_i | k).
\end{equation}
This mean likelihood \( \bar{p}(k) \) provides an aggregate measure of how representative Gaussian \( k \) is for the examples of the entire bag. After computing \( \bar{p}(k) \) for each Gaussian \( k \), the final representation \( \mathbf{r} \) of the bag is:
\begin{equation}
\mathbf{r} = [\bar{p}(1), \bar{p}(2), \ldots, \bar{p}(K)] \in \mathbb{R}^K.
\end{equation}
In this case, \( \mathbf{r} \) encapsulates the averaged likelihoods for each Gaussian, providing a compact invariant representation of the bag in the latent space. The size of this representation will be equal to the number of Gaussian distributions $K$ used to model the latent space.

It is important to note that we can expand this layer by working with a number $L$ of different latent spaces expanded in width (see Figure~\ref{fig:architecturehorizontal}). The full representation \( \mathbf{r} \) of the bag is computed by concatenating the representations from all latent spaces: 
\begin{equation}
\mathbf{r} = [\mathbf{r}_1, \mathbf{r}_2, \ldots, \mathbf{r}_L] \in \mathbb{R}^{LK}.
\end{equation}
Here, \( \mathbf{r} \) combines all individual representations into a single vector, providing a comprehensive representation of the bag across multiple latent spaces. In this particular case, the FEM is replicated across every latent space, to make the projection unique and different for each of them (see Figure~\ref{fig:architecturehorizontal}).

\begin{figure}[t]
    \centering
    
    % Subfigure 1
    \begin{subfigure}[b]{0.32\textwidth}
        \centering
        \includegraphics[width=\textwidth]{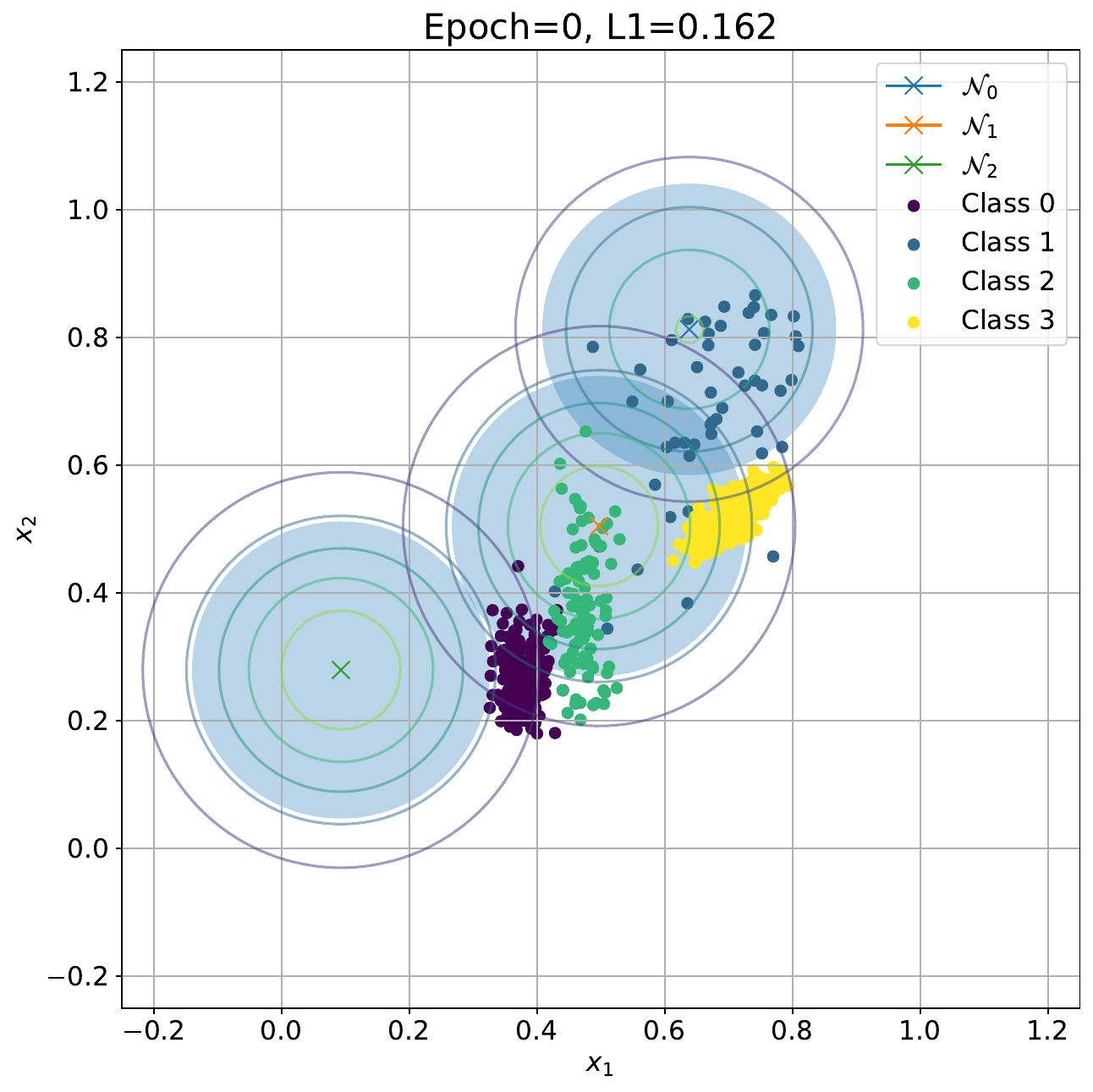}
        \label{fig:sub1}
    \end{subfigure}
    \hfill
    % Subfigure 2
    \begin{subfigure}[b]{0.32\textwidth}
        \centering
        \includegraphics[width=\textwidth]{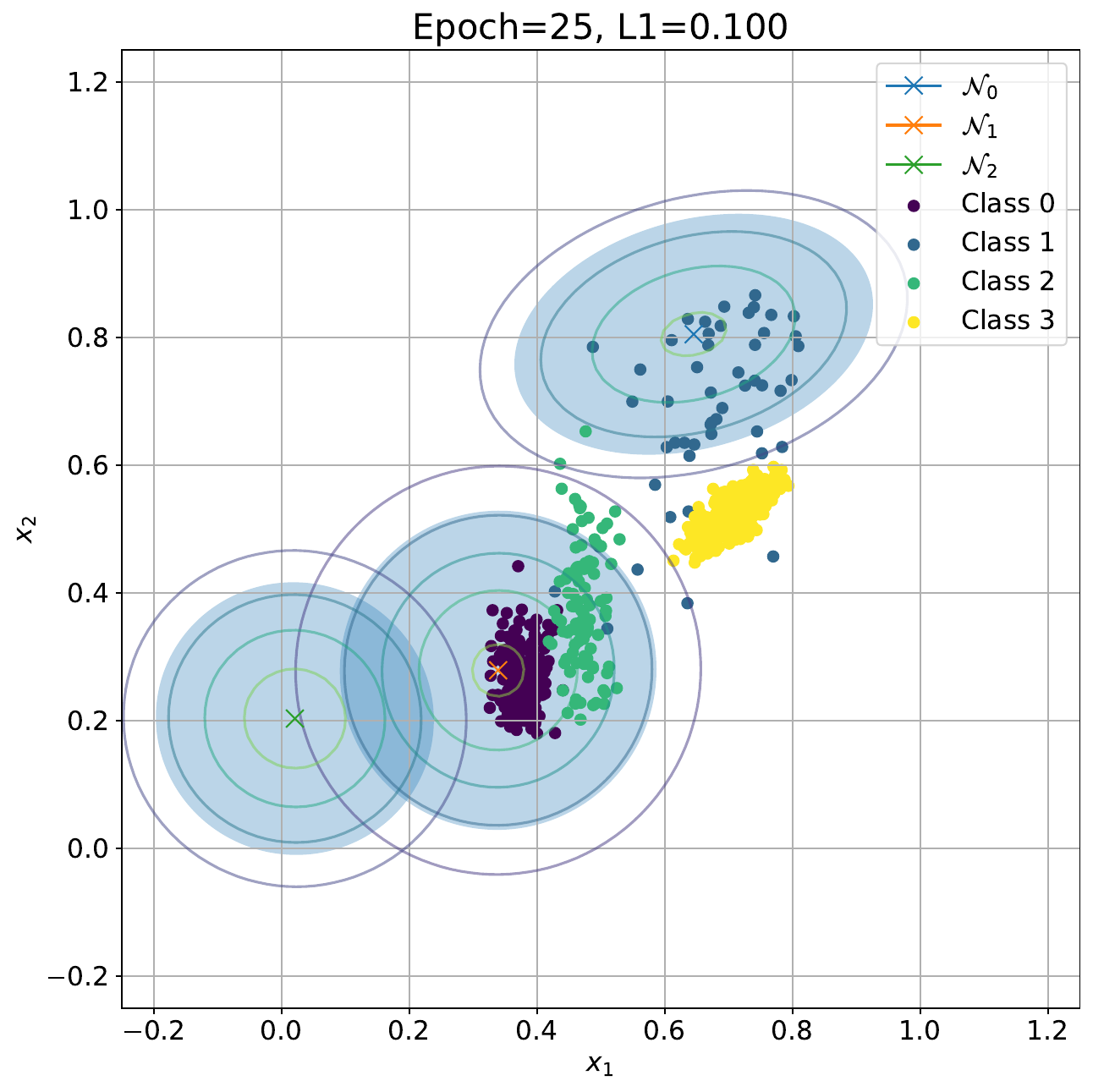}
        \label{fig:sub2}
    \end{subfigure}
    \hfill
    % Subfigure 3
    \begin{subfigure}[b]{0.32\textwidth}
        \centering
        \includegraphics[width=\textwidth]{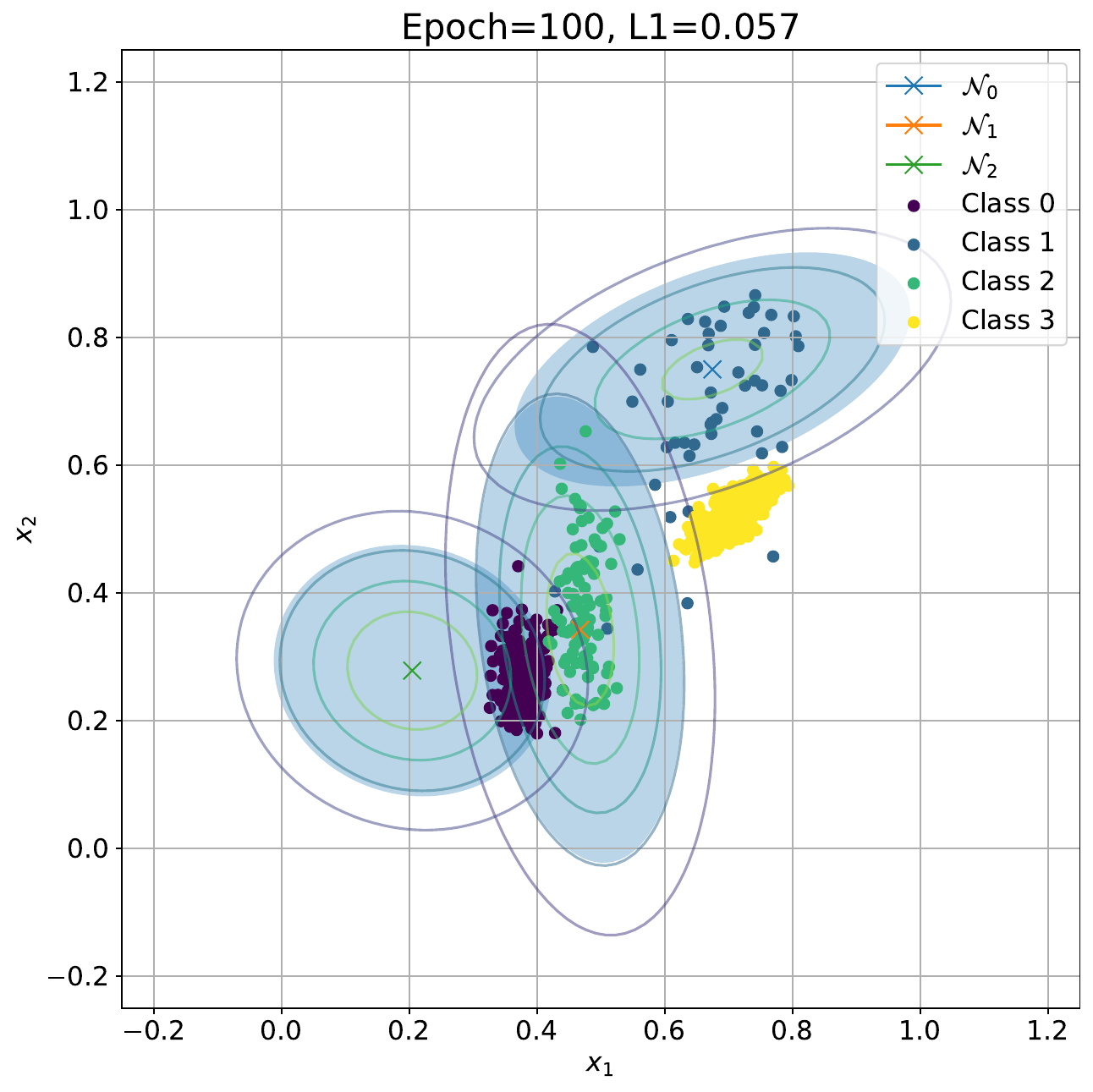}
        \label{fig:sub3}
    \end{subfigure}

    \caption{Example of the BRM inner workings using a bare-bones architecture. Input data corresponds to a synthetic multiclass problem with four classes and two input dimensions. Data is projected to a latent space just using a \textit{sigmoid} function (note that in the full network, the FEM will be in charge of projecting input data to the latent space). For the easiness of visualization, just three Gaussian distributions are in charge of modeling input data. As training progresses, the network learns to place the Gaussian distributions 
    in convenient places to create a useful representation $\mathbf{r}$ of the bag, that the QM can use with the objective of minimizing the chosen loss function (L1 in this case). Note that the number of classes and the number of Gaussian distributions 
    are not directly related, as the network will learn their centers and shapes to optimally represent the data (in the experiments, we will use 100 Gaussians \added{for problems with up to 28 classes).}}
    \label{fig:examplegaussians}
\end{figure}

This bag representation $\mathbf{r}$ is then used by the QM (see Figure~\ref{fig:architecture}) to predict a final prevalence vector for the bag in question. As training progresses the network will back-propagate the loss function (a quantification loss) and learn to place and shape the Gaussian distributions so the BRM is capable of providing a good representation for the QM to predict good prevalence estimates (see Figure~\ref{fig:examplegaussians}).

The difference between this BRM and those based on basic pooling functions like \textit{average}, \textit{max}, or \textit{median} is substantial, as it provides a more expressive representation than merely computing a single statistic per feature. While the network can adapt its layers to leverage the available information, providing a better embedding for the bag aids significantly in converging to an optimal solution. The distinction from a histogram layer is more subtle. Firstly, histograms are inherently discrete, whereas Gaussian distributions offer a continuous representation that avoids binning. Secondly, creating a histogram per feature in the latent space loses the connections between features, whereas Gaussian distributions may capture these correlations by considering all features simultaneously, resulting in a richer and more accurate representation.

\subsection{Parameter initialization}
Parameter initialization is a key ingredient for successfully training a neural network. In the architecture presented, each representation layer needs to place a set of $K$ Gaussian distributions in the latent space. For each Gaussian distribution, its center $\boldsymbol{\mu}_i$ and its covariance matrix $\mathbf{\Sigma}_i$ are network parameters that require initialization. 

The BRM is preceded by a \textit{sigmoid} function, which confines the latent space to $[0, 1]$, resembling a hypercube of dimension $d$. Our design choice was to initialize the Gaussian centers $\boldsymbol{\mu}_i$ uniformly distributed across all dimensions within the range $[0,1]$. 

Regarding the initial covariance matrix \(\boldsymbol{\Sigma}_{i}\), we adopt a diagonal matrix of size $d$ (the dimensionality of the latent space). The diagonal elements $\boldsymbol{\Sigma}_{ii}$ are initialized based on a heuristic involving pairwise distances between Gaussian centers $\boldsymbol{\mu}_{i}$. Specifically, we compute these distances (excluding self-distances) and use the average minimum distance divided by 2, squared:
\begin{equation}
\boldsymbol{\Sigma}_{ii} = \left( \frac{\text{mean}\left(\text{min}_{i \neq j} |\boldsymbol{\mu}_i - \boldsymbol{\mu}_j|\right)}{2} \right)^2.
\end{equation}
The intuition behind this heuristic is to make the Gaussians cover the maximum space possible in the hypercube without excessive overlapping. 

As each element of the covariance matrix $\boldsymbol{\Sigma}_{i}$ serves as a parameter in the network, it is crucial to ensure that $\mathbf{\Sigma}_{i}$ remains a valid covariance matrix, meaning it must be positive-definite during all the training process. To achieve this, we utilize \textit{geotorch} \citep{lezcano2019trivializations}, a library that enables us to enforce the positive-definiteness of each $\mathbf{\Sigma}_{i}$. This constraint is essential for maintaining the integrity of Gaussian distributions in our model, supporting robust training and reliable inference.

The rest of the network parameters (linear layers, batch norm layers, etc) are initialized with the default procedure used in the PyTorch deep learning framework.

\subsection{Latent space similarity regularization}
\label{sec:regularization}
As explained in Section~\ref{sec:latentrepr}, our network architecture is extended horizontally to incorporate multiple latent space representations, aiming to capture diverse information about each bag. Each latent representation operates with its own set of Gaussian distributions to model and derive a robust invariant representation essential for quantification tasks.

The network by itself, when trying to optimize for the loss provided, will try to make use of these different latent space representations to make a powerful joint representation. A way of helping the network in this process could be enforcing that the latent spaces differ among them. To compute the similarity between latent spaces we have used the score provided by Centered Kernel Alignment (CKA) \citep{kornblith2019similarity}, which is able to compute the similarity of latent spaces of any dimension. 

The CKA score measures the similarity between representations learned by different neural networks even with different dimensions provided that they are evaluated with the same set of examples. Formally, given a set of $L$ latent space activations \( \{\mathbf{Z}_1 \in \mathbb{R}^{n \times d_1}, \mathbf{Z}_2 \in \mathbb{R}^{n \times d_2}, \ldots, \mathbf{Z}_L\ \in \mathbb{R}^{n \times d_L}\} \), where $n$ is the number of examples and $d$ the dimension of each latent space, the CKA score is computed as follows:
\begin{equation}
\text{CKA}(\{\mathbf{Z}_i\}) = \frac{1}{\binom{L}{2}} \sum_{i < j} \frac{\|\mathbf{Z}_i^\top \mathbf{Z}_j \|_F^2}{\|\mathbf{Z}^\top_i \mathbf{Z}_i\|_F \|\mathbf{Z}^\top_j \mathbf{Z}_j\|_F},
\end{equation}
where $\|\cdot \|_F$ denotes the Frobenius norm. The numerator computes the similarity between representations $\mathbf{Z}_i$ and $\mathbf{Z}_j$ while the denominator serves to normalize the similarity to ensure it is not influenced by the scale of the activations, thereby providing a scale-invariant measure of alignment between the latent spaces. 

By incorporating this similarity metric into the loss function with an appropriate weighting factor (a hyperparameter), the network is compelled to converge towards solutions where the latent spaces exhibit greater differentiation and consequently offer more informative representations. Our experimental findings show that integrating this approach sometimes improves not only the results but also accelerates network convergence.

The CKA regularization term is added to the total loss of the network before backpropagation, multiplied by a regularization factor \( \lambda \). The modified loss function is:
\begin{equation}
\mathcal{L}_{\text{total}} = \mathcal{L}_{\text{original}} + \lambda \cdot \text{CKA}(\{\mathbf{Z}_i\}),
\end{equation}
where \( \mathcal{L}_{\text{original}} \) is the original loss function of the network (usually a quantification loss), \( \lambda \) is the regularization factor, and \( \text{CKA}(\{\mathbf{Z}_i\}) \) is the Centered Kernel Alignment score computed over the set of latent space projections \( \{\mathbf{Z}_i\} \).

\subsection{Data augmentation}
\label{sec:dataaugmentation}
Every deep learning system is highly dependent on the amount of data available. For quantification, the ideal scenario involves having bags of examples labeled by prevalence. In a symmetric setting, even with a large number of labeled bags, each bag in quantification becomes a single training example, which makes typical dataset sizes relatively small for deep learning. For instance, in our experiments, one thousand labeled bags by prevalence represent a low number of training examples for a neural network, making data augmentation techniques essential.

In the context of quantification, data augmentation can be achieved by mixing bags to create new ones \citep{perez2024quantification}. This approach is particularly useful when the number of labeled bags is limited. By randomly picking and mixing two real bags, we can produce a new \textit{augmented} bag. The prevalence of the new bag is computed as the mean class prevalences of the two mixed bags. This data augmentation technique is called "Bag Mixer" and will be used in the experiments for all deep learning methods.

It is important to note that when individual example labels are available alongside bags labeled by prevalence, they can be leveraged to generate additional training \added{bags}. In this scenario, we can generate new synthetic training bags with dataset shift for training deep quantifiers. For instance, to create new bags with prior probability shift, we can just make a grid and change the class priors of each class step by step. While this procedure might work well for two classes, it gets unfeasible when the number of classes increases. The alternative is to use an algorithm like Kraemer \citep{smith2004sampling} to draw uniform
prevalence vectors at random and then obtain the bags matching these prevalence values using sampling with
replacement from the training dataset. This protocol for generating bags is usually known in the quantification
literature as APP protocol \citep{quantbook2023}.  As we will demonstrate in the experiments and in consonance with previous research in deep learning, the more amount and variety of data provided to the network, the better its performance and convergence (a result that, on the other hand, was expected). 

\section{Experiments}
\label{sec:experiments}

\subsection{Datasets}
All the datasets included in the experiments are part of the only two quantification competitions held up to date (LeQua, editions 2022 \citep{lequa2022} and 2024\footnote{\url{https://lequa2024.github.io}}). The chosen datasets for this study correspond to multiclass (with names T1B for 2022 and T2 for 2024) and ordinal quantification (with name T3), as they are the more challenging and the ones that pose bigger difficulties to quantification methods. All datasets are freely available for download and consist of product reviews extracted from Amazon. 

For multiclass quantification tasks (T1B and T2), each review belongs to one of 28 different merchandise classes (“Automotive”, “Baby”, “Beauty”, ...). Our task is to predict the class prevalences of bags of reviews affected by prior probability shift. The only difference between T1B and T2 is the input space of the features. T1B provides 300 numeric features per review while T2 provides only 256. The objective function for these tasks is Relative Absolute Error (RAE), defined as:
\begin{equation}
    \label{eq:rae}
    \text{RAE}(\mathbf{p},\hat{\mathbf{p}})=\frac{1}{|\mathcal{Y}|}\sum_{c_i\in\mathcal{Y}}\frac{|\delta(p(c_i))-\delta(\hat{p}(c_i))|}{\delta(p(c_i))}, 
\end{equation}
in which $\delta(p_i) = \frac{p_i+\epsilon}{|\mathcal{Y}|\epsilon+1}$ is the smoothing function, with $\epsilon$ the smoothing factor that
is set to $(2|B|)-1$ following \citet{Sebastiani2020}. 

% A second loss function taken into account to evaluate multiclass quantifiers is the absolute error, computed as:

% \begin{equation}
%     \label{eq:ae}
%     AE(\mathbf{p},\hat{\mathbf{p}})=\frac{1}{|\mathcal{Y}|}\sum_{c_i\in\mathcal{Y}}|p(c_i)-\hat{p}(c_i)|.
% \end{equation}

The ordinal quantification task consists on predicting the prevalence of each class, in this case, sentiment-based polarity, expressed as a star rating (from "1 star" to "5 stars") in bags of product reviews. The particularity of this task is that classes have implicit order which means that mistakes do not count the same when predicting the class prevalences. This is well expressed by the loss function, Normalized Matched Distance (NMD), used in this task:
\begin{equation}
\text{NMD}(\mathbf{p}, \hat{\mathbf{p}}) = \frac{1}{|\mathcal{Y}|-1}\sum_{j=1}^{|\mathcal{Y}|-1}  |\sum_{i=1}^j \hat{p}(c_i)-\sum_{i=1}^j p(c_i)|,
\end{equation}
where $\frac{1}{|\mathcal{Y}-1|}$ is a normalization term so NMD ranges between 0 (best) and 1 (worst). NMD is designed to guide the optimization process in ordinal quantification tasks by quantifying how well the predicted prevalences match the true prevalences in a normalized and class-aware manner, where class order is taken into account.

All three datasets provide 20,000 individual labeled examples, along with 1,000 bags labeled by prevalence. Each training bag contains 1,000 individuals for T1B and T2, and 200 individuals for T3. It is important to note that these bags are labeled by prevalence but the label of each example in them is not available. The test set comprises 5,000 bags for each task, each containing 1,000 elements for T1B and T2, and 200 for T3.

%LeQua2022-T1B is a multiclass task focused on estimating the prevalence of 28 merchandise product categories. For this task, the organizers provided a training set comprising 20,000 labeled opinions represented by 300 features. Additionally, a validation set consisting of 1,000 bags, each containing 1,000 unlabeled documents, annotated by prevalence was provided. The testing set included 5,000 bags, each with 1,000 documents.

%In the case of LeQua2024-T2, it is the analogous task to LeQua2022-T1B in LeQua competition 2024. The difference between them is the number of features used to represent the reviews, in T2 it is 256. As in the previous task, LeQua2024-T2 includes a training set with 20,000 labeled reviews, a validation set with 1,000 bags of 1,000 unlabeled opinions annotated by prevalence, and a testing set comprising 5,000 bags of 1,000 documents each.

%Regarding LeQua2024-T3, this task is a multiclass-ordinal task that contains sets of reviews labeled by sentiment-based polarity, expressed as star ratings (from 1 to 5 stars). The provided datasets include a training set with 100 bags of 300 labeled reviews, annotated by prevalence; a validation set with 1,000 bags of 300 unlabeled opinions, also annotated by prevalence, and a testing set of 5,000 bags, each containing 300 reviews.

\subsection{Experiments in the LeQua datasets}
\label{sec:explequa}
This section describes the experiments in the three LeQua datasets. Two main approaches were employed: traditional quantifiers, trained with the QuaPy public quantification library \citep{moreo2021quapy}, and quantification methods based on deep learning. On the one hand, traditional quantifiers, as detailed in Section~\ref{sec:related}, require a classifier that was trained using datasets with individual example labels. This means that training bags labeled by prevalence can not be used in the training process directly. Instead, they are used to perform a grid search to optimize classifier and quantifier hyperparameters, aiming for the best quantification performance based on official evaluation metrics (RAE for T1B and T2, and NMD for T3). On the other hand, quantification methods based on deep learning are optimized based on a chosen loss function (as before, RAE, for T1B and T2 and NMD for T3). 

For the three datasets, two different setups were considered for the deep learning lot:
\begin{itemize}
    \item \textbf{U (unlabeled) setting}. In this scenario, deep learning methods use only provided bags for training, discarding the labeled dataset. 70\% of the available bags labeled by prevalence (700 bags) were used for training, while the other 30\% (300 bags) were used for validation and early stopping.
    \item \textbf{U+APP setting}. In this scenario, apart from 700 training bags, we feed the network with extra synthetically generated training bags presenting prior probability shift, using the APP protocol (see Section~\ref{sec:dataaugmentation}) over the individually labeled examples. In this setup, APP-generated bags represent the 50\% of all the bags fed to the network.
\end{itemize}
In both settings, bag-labeled data was augmented by randomly mixing real bags using the \textit{Bag Mixer} described in Section~\ref{sec:dataaugmentation}. Running the experiments without the \textit{Bag Mixer} was discarded because as the ablation study in \cite{perez2024quantification} demonstrates that deep learning quantification methods suffer significantly from overfitting without this data augmentation strategy. %Additionally, we conducted experiments using only the dataset labeled at the individual example level, where synthetic bags were generated via the APP protocol, and bags labeled by prevalence were used only for validation purposes. However, this setup inherently provides limited data for training deep neural networks, which traditionally require large datasets to perform effectively. Consequently, the performance in this scenario is expected to be suboptimal.}

While the hyperparameters of traditional quantification were optimized using the validation dataset, an exhaustive hyperparameter search for deep learning methods was not feasible due to limited resources. The size of the networks was adjusted to fit within a single small GPU (with 12 GB of RAM). All deep learning methods were trained using the same network architecture, varying only the BRM, to ensure a fair comparison between them. In the BRM module we tested basic pooling layers —such as average, median, and max— referred to as DQN \citep{Qi2021}; differentiable histograms, named HistNetQ \citep{perez2024quantification}; and the layer proposed in this paper based on multivariate Gaussian distributions, called GMNet (see Section~\ref{sec:latentrepr}). The number of Gaussian distributions was set to 100 for each latent space, with each space in 5 dimensions. We used 9 latent spaces. The value for the CKA regularization was fixed to 0.01 for all experiments. For HistNetQ, we used a histogram with 32 bins per feature.

The FEM utilized in these experiments comprises a series of fully-connected layers with dropout, given that all three datasets used consist of tabular data. The last layer of the FEM, which corresponds to the input of the BRM had a size of 512. It should be noted that in GMNet, with latent spaces, the FEM is integrated within each submodule to obtain different projections (see Figure~\ref{fig:architecturehorizontal}). In this case, we chose an output size for each FEM in each submodule of 50 neurons. 
In terms of the QM, all networks use the same set of fully connected layers with dropout followed by a softmax activation function, which will finally output the estimated prevalence of the classes.

All networks were trained using an identical procedure to ensure comparability. \footnote{Code and data to fully reproduce the experiments can be found in \url{https://github.com/AICGijon/gmnet}.} We applied early stopping after 40 epochs in which the validation loss did not improve.  

\begin{table*}[t]
  \caption{Results for \textsc{T1B}, \textsc{T2} and \textsc{T3}, in terms of RAE (for \textsc{T1B} and \textsc{T2}) and NMD (for \textsc{T3}). Deep learning methods where training using only bags labeled by prevalence (U) and bags labeled by prevalence plus synthetic bags generated using the APP protocol (U+APP).}
  \label{tab:results}
  \centering
  \begin{tabular}{r|ccc}
   \toprule
   &T1B (RAE) & T2 (RAE) & T3 (NMD)\\
   \midrule
   CC  	    	 	& 1.8936 $\pm$ 1.187 & 2.3096 $\pm$ 1.384	& 0.0809 $\pm$ 0.047\\
   PCC 	      	& 2.2646 $\pm$ 1.416 & 2.6488 $\pm$ 1.610 & 0.0669 $\pm$ 0.051\\
   ACC      	 	& 1.4213 $\pm$ 1.270 & 1.3480 $\pm$ 1.164 & 0.1193 $\pm$ 0.065\\
   PACC     	 	& 1.3054 $\pm$ 0.988 & 1.1956 $\pm$ 1.137 & 0.1246 $\pm$ 0.066\\
   DMy      		& 1.0419 $\pm$ 1.030 & 1.3189 $\pm$ 1.128 &	0.1156 $\pm$ 0.065\\
   EMQ-NoCalib 	& 0.8780 $\pm$ 0.751 & 1.1623 $\pm$ 0.991 &	0.1130 $\pm$ 0.070\\
   EMQ-Platt 		& 1.1959 $\pm$ 1.137 & 1.9203 $\pm$ 1.900 &	0.1806 $\pm$ 0.099\\
   \hline
    DQN (avg) - U & 0.9586 $\pm$ 0.629 & 1.3169 $\pm$ 0.936 & 0.0552 $\pm$ 0.041 \\
    DQN (max) - U & 1.3163 $\pm$ 0.882 & 2.0445 $\pm$ 1.357 & 0.2247 $\pm$ 0.060 \\
    DQN (med) - U & 0.7893 $\pm$ 0.499 & 1.1779 $\pm$ 0.847 & 0.0546 $\pm$ 0.040 \\
    HistNetQ - U & 0.6989 $\pm$ 0.451 & 0.9162 $\pm$ 0.598 & 0.0489 $\pm$ 0.036 \\
    GMNet (ours) - U & 0.6728 $\pm$ 0.368 & 0.8816 $\pm$ 0.550 & 0.0502 $\pm$ 0.037 \\
    \hline
    DQN (avg) - U+APP & 0.7970 $\pm$ 0.524 & 1.3105 $\pm$ 0.968 & 0.0537 $\pm$ 0.040 \\
    DQN (max) - U+APP & 0.9211 $\pm$ 0.571 & 2.4288 $\pm$ 1.729 & 0.0587 $\pm$ 0.043 \\
    DQN (med) - U+APP & 0.7286 $\pm$ 0.466 & 1.0852 $\pm$ 0.815 & 0.0534 $\pm$ 0.039 \\
    HistNetQ - U+APP & 0.5711 $\pm$ 0.374 & 0.7455 $\pm$ 0.501 & \bftab 0.0467 $\pm$ 0.036 \\
    GMNet (ours) - U+APP & \bftab 0.5433 $\pm$ 0.263 & \bftab 0.7062 $\pm$ 0.420 & 0.0498 $\pm$ 0.037 \\
    % \hline
    % DQN(avg) - APP & 1.4938 $\pm$ 1.748 & 1.6715 $\pm$ 2.103 & 0.0953 $\pm$ 0.063 \\
    % DQN(max) - APP & 2.9695 $\pm$ 3.958 & 2.9737 $\pm$ 2.764 & 0.1189 $\pm$ 0.071 \\
    % DQN(med) - APP & 1.2518 $\pm$ 1.145 & 1.5940 $\pm$ 1.780 & 0.1199 $\pm$ 0.057 \\
    % HistNetQ - APP & 2.2146 $\pm$ 2.859 & 2.0289 $\pm$ 2.290 & 0.0857 $\pm$ 0.058 \\
    % GMNet (ours) - APP & 1.2921 $\pm$ 1.297 & 1.5871 $\pm$ 1.652 & 0.1051 $\pm$ 0.073 \\
   \bottomrule
 \end{tabular}
\end{table*}

In Table~\ref{tab:results}, we present the performance of the quantification methods on the three datasets.
%: T1B, T2, and T3. On the one hand, we evaluate two multiclass datasets with 28 classes (T1B and T2). Both represent the same problem setting but with different features and different data. However, T2 is more challenging than T1B, as it has a smaller number of features and of lower quality. On the other hand, T3 is a multiclass ordinal problem with five classes.
A detailed analysis of the results for the T1B dataset reveals that GMNet outperforms all other methods, including traditional and deep learning-based ones, in both the U and U+APP settings. This is particularly notable as GMNet significantly surpasses HistNetQ, the previously best-known method for this quantification task \citep{perez2024quantification}. The difference between U+APP and U setups is notable as all deep learning methods improve their results when using also bags generated with APP. This underscores how deep learning methods benefit from additional training data. Among the deep learning methods, the ones with more complex BRMs (histograms and Gaussian distributions) stand out as the best compared to methods with simpler representations as median or average, that fall back but perform exceedingly well compared with traditional methods. Among traditional methods, EMQ without calibration stands out as the best performer. These results highlight the advantage of deep learning methods in optimizing specific loss functions and the ability to train directly using bags, compared to traditional quantification methods, which do not have these capabilities.

Concerning the T2 dataset, the same tendencies are observed as for T1B. GMNet comes out as the best performer and the best deep learning methods (GMNet, HistnetQ, and DQN (med)) outperform by a wide margin all traditional methods, for the optimized loss (RAE), including EMQ, which is considered one of the best quantification methods in the literature \citep{alexandari2020maximum}. When the amount of data is reduced (U setting) the performance of the deep learning methods systematically decreases.

In the T3 dataset, deep learning methods once again demonstrate superior performance, but now with HistNetQ slightly ahead of GMNet. The capability to effectively optimize loss functions designed for ordinal quantification further strengthens the advantage of these methods in tackling such tasks. It is interesting to note that in this dataset, even the BRM that consistently gets the worst results among the deep learning lot (\emph{max} pooling layer), outperforms all traditional methods in the U+APP setting.

\subsection{Impact of data availability on quantification methods}
The previous section demonstrates that, in the context of the LeQua quantification competition, there exists a substantial performance gap between the best deep learning methods (HistNetQ and GMNet) and traditional quantification methods in the three tasks analyzed. However, to better understand the underlying reasons for this disparity, we designed an additional experiment aimed at answering, in our opinion, two key questions: i) Is the gap primarily due to the inability of traditional methods to exploit the same training data as deep learning methods? ii) How does the amount of training data affect the performance of deep learning methods in quantification tasks?

\begin{figure}[t]
  \centering
    \includegraphics[width=\textwidth]{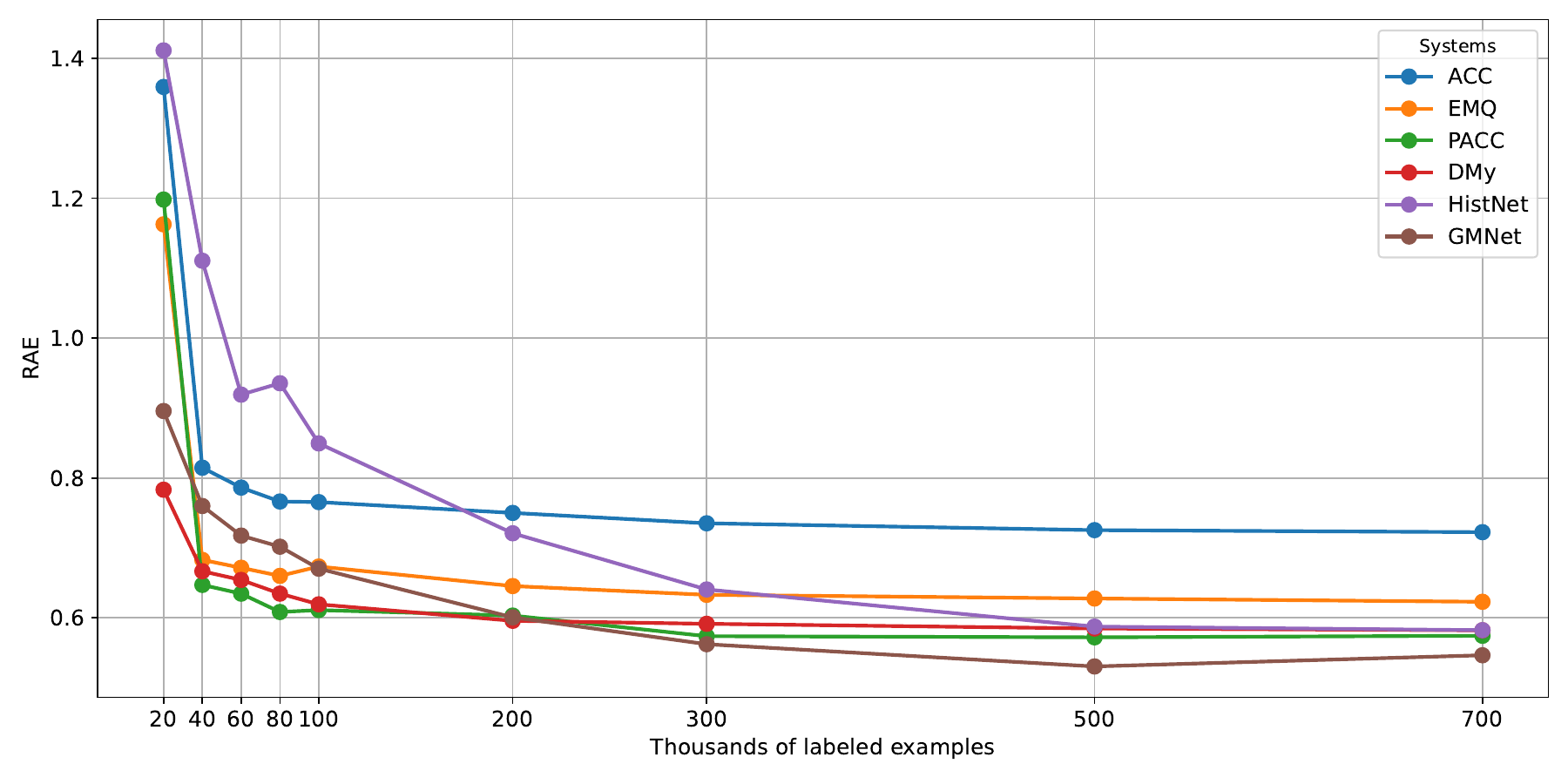}
    \caption{U+APP setting over the T2 dataset.}
    \label{fig:resultsapp_u}
\end{figure}

In the main experiment of the paper (see Section~\ref{sec:explequa}), deep learning methods are able to exploit bags labeled by prevalence for training, a capability that traditional methods lack (although, they use this data for validation and hyperparameter tuning). To perform this experiment, we requested the individual labels of the 1000 bags labeled by prevalence of dataset T2 to the organizers of the quantification competition LeQua2024, enabling traditional quantifiers to train using this data. This new dataset comprises 1 million labeled examples. We conducted experiments with different training-validation splits incrementally, from 20k to 700k training examples. The same two settings as in the main experiment were considered for the deep learning methods (U and U+APP). In the U setting only bags labeled by prevalence are used for training deep learning methods while traditional methods will use all the examples in these bags along with their labels. In the U+APP setting, the label of the examples in the training bags is used also for generating new synthetic training bags with prior probability shift (using the APP protocol) that are fed to the network along with the training bags. 

\begin{figure}[t]
    \centering
    \includegraphics[width=\textwidth]{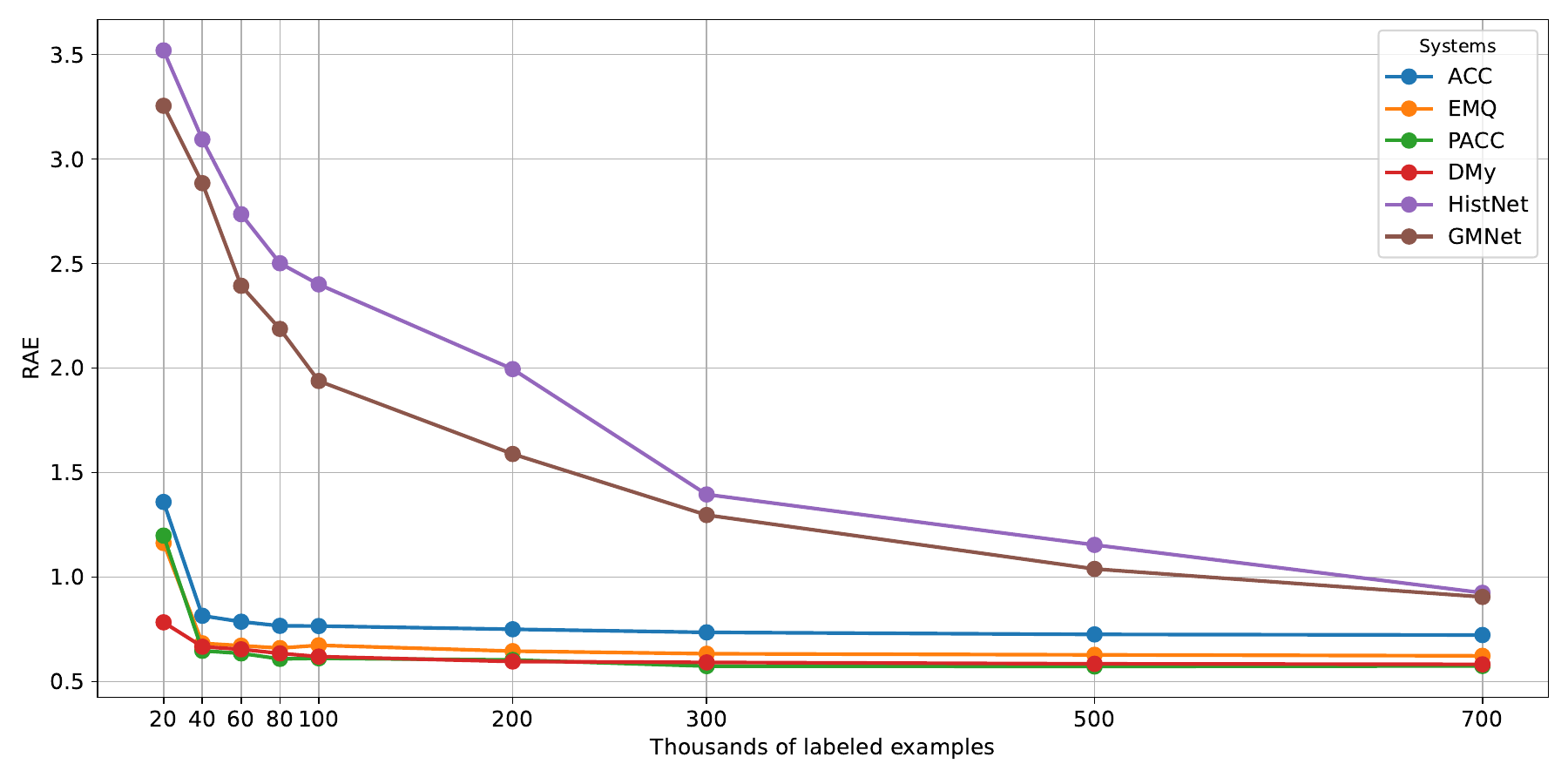}
    \caption{U setting over the T2 dataset.}
    \label{fig:results_u}
\end{figure}

Figure~\ref{fig:resultsapp_u} presents the results for the U+APP setting, while Figure~\ref{fig:results_u} shows the results for the U setting on the T2 dataset using different training-validation splits. For clarity, only the best-performing methods of each type are included in these experiments. 

As expected, results show that traditional methods benefit significantly from having more than 20k labeled examples. Training a classifier to estimate probabilities in a 28-class problem is highly challenging with only 20k examples (i.e. $\sim$710 examples per class). Results show that with 100k examples traditional methods obtain already very good results, with only marginal improvements beyond this point.

On the other hand, deep learning methods, such as HistNetQ and GMNet, tend to perform better as the amount of data increases, a trend expected when using deep learning. The results indicate that GMNet, using the U+APP setting, becomes competitive with the best traditional quantification methods when trained on 200k examples (200 training bags), with its performance continuing to improve as the dataset size increases. Conversely, HistNetQ needs more training data to be competitive,  not matching the best traditional methods until 500 training bags are provided. In the U setting (see Figure~\ref{fig:results_u}), deep learning methods only use a very limited number of training bags (augmented with the \emph{Bag Mixer}). Even with 700 training bags, the results obtained are not comparable to those achieved by traditional methods using 700k labeled examples. For instance, when using 20k examples, we are only providing the network with 20 different bags (each bag has 1000 examples), while attempting to estimate prevalences in a 28-class problem, making the task almost impossible. Note that APP (used in the U+APP setting) greatly helps the network introducing a lot of variability in the training bags, that are synthetically created with different examples (picked at random) and different prevalences (generated using the Kraemer algorithm), covering the entire prevalence space. APP is the only mechanism by which these systems leverage individual example labels. In the U setting, only the available training bags are fed to the network. This setup is disadvantageous for the networks as the amount of training instances is very limited.

In Table~\ref{tab:results}, HistNetQ results may appear quite similar to those of GMNet. This experiment highlights that GMNet exhibits a reduced dependence on data volume compared to HistNetQ, remaining competitive even with limited data. The performance gap between the two models is substantial when using small amounts of training data but narrows as more data becomes available to the networks.  %Note that this experiment differs from the main one in that here we only use the training bags corresponding to the labeled examples, whereas in the main experiment, we always used 700 bags labeled by prevalence for training.

\subsection{Latent space regularization}

To evaluate how latent space regularization (see Section~\ref{sec:regularization}) affects GMNet, we conducted various experiments applying different values of CKA regularization ($\lambda$) in all three datasets, keeping thee rest of the training hyperparameters untouched. Their performances are presented in Table ~\ref{tab:results2}. On the T1B and T2 datasets, CKA regularization enhances performance to some degree, while on T3, the tested values of $\lambda$ yield no observable improvement. In this case, the BRM may play a less critical role, as even a simple pooling mechanism like \emph{max} achieves competitive results.

In summary, these results show that regularization may help improve performance, helping the network to converge to solutions with different latent spaces that provide richer information. As with any other hyperparameter, the correct value must be set for each problem. Even though the margin is not wide, it usually provides faster convergence time that pays off the extra amount of computation needed for calculating the CKA term.

\begin{table*}[t]
 \caption{Results for GMNet using different $\lambda$ values controlling the CKA regularization. }
 \label{tab:results2}
 \centering
 \begin{tabular}{l|cccc}
	\toprule
	$\lambda$ & T1B (RAE) & T2 (RAE) & T3 (NMD) \\
	\midrule
    0 & 0.5531 $\pm$ 0.272 & 0.7130 $\pm$ 0.392 & \bftab 0.0485 $\pm$ 0.036 \\
    0.1 & \bftab 0.5307 $\pm$ 0.286 & 0.7113 $\pm$ 0.361 & 0.0507 $\pm$ 0.038 \\
    0.01 & 0.5433 $\pm$ 0.263 & \bftab 0.7062 $\pm$ 0.420 & 0.0498 $\pm$ 0.037 \\
    0.001 & 0.5345 $\pm$ 0.288 & 0.7113 $\pm$ 0.388 & 0.0492 $\pm$ 0.035 \\
	\bottomrule
\end{tabular}
\end{table*}

\section{Conclusions}
\label{sec:conclusions}

In this paper, we have presented a neural network for quantification that utilizes Gaussian distributions in latent spaces to obtain invariant representations of bags of examples, called GMNet. This representation proved useful to capture relationships among features in complex multiclass quantification problems, providing state-of-the-art results in the T1B and T2 datasets, the standard benchmarks for the quantification community in the multiclass setting. Additionally, the inherent ability of deep learning methods to train directly on bags and optimize task-specific loss functions offers significant advantages over traditional quantification methods, which lack these capabilities. This feature is particularly beneficial in problems with highly specific loss functions, such as T3, an ordinal quantification task where the NMD loss function is employed.

These outstanding results highlight some of the advantages of the proposed architecture, GMNet, including: i) the quality of the proposed bag representation module (BRM), which is able to capture complex iterations between features projected in latent spaces, that improves the results obtained by previous methods; ii) the capability to directly optimize a specific loss function, as demonstrated in T1B and T2 by optimizing RAE, and in T3 by optimizing the ordinal loss NMD; iii) the ability to perform without example-labeled data,  which is an interesting capability that traditional quantification methods do not have; iv) its performance with limited data compared to other deep learning quantification methods (HistNetQ); v) the effectiveness of regularization in complex problems, allowing GMNet to solve them more efficiently.

Future work may include extending this study to other datasets, which have proven particularly challenging for traditional quantification methods \citep{gonzalez2019automatic} and where training bags are naturally present. Also, the application of these methods to problems where other types of shift different from prior probability shift occur \citep{gonzalez2024binary}. Additionally, our BRM has potential applications beyond quantification, such as in general set processing tasks \citep{zaheer2017deep}, where permutation invariant representations are essential, or Learning from Label Proportions (LLP) \citep{yu2014learning}. A final avenue for future work, and maybe the most critical to improving the proposed method, could focus on how to make better use of the data available, as for instance new data augmentation techniques or other forms of synthetic training bag generation.
\\
\section*{Acknowledgments}
This work has been funded by Agencia de Ciencia, Competitividad Empresarial e Innovación Asturiana (Sekuens) within the program "Grants for research groups from organizations in the Principado de Asturias, 2024" (GRU-GIC-24-018).

The authors would like to thank Dr. Alejandro Moreo and Dr. Pablo Pérez for their insightful discussions about the paper.

\bibliographystyle{unsrtnat} 
\bibliography{gmnet}  

\end{document}